\documentclass[
]{ceurart}

\usepackage{xcolor}
\usepackage{graphicx}
\usepackage{caption}
\usepackage{subcaption}
\usepackage{float}
\usepackage{enumitem}
\usepackage{listings}
\begin{document}

\copyrightyear{2024}
\copyrightclause{Copyright for this paper by its authors.
  Use permitted under Creative Commons License Attribution 4.0
  International (CC BY 4.0).}

\conference{HCI SI 2024: Human-Computer Interaction Slovenia 2024, November 8th, 2024, Ljubljana, Slovenia}

\title{Teaching Shortest Path Algorithms With a Robot and Overlaid Projections}


\author[1]{Pavel Jolakoski}[%
orcid=0009-0006-4515-5390,
email= pavel.jolakoski@gmail.com
]
\author[1,2]{Jordan Aiko {Deja}}[%
orcid=0000-0001-9341-6088,
email=	jordan.deja@famnit.upr.si
]
\author[1,3,4]{Klen {Čopič Pucihar}}[%
orcid=0000-0002-7784-1356, 
email=klen.copic@famnit.upr.si,
]
\author[1,4]{Matjaž Kljun}[%
orcid=0000-0002-6988-3046,
email=matjaz.kljun@upr.si,
]
\address[1]{University of Primorska, Faculty of Mathematics, Natural Sciences and Information Technologies, Koper, Slovenia}
\address[2]{De La Salle University, Manila, Philippines}
\address[3]{Faculty of Information Studies, Novo Mesto, Slovenia}
\address[4]{Stellenbosch University, Department of Information Science, Stellenbosch, South Africa}

\begin{abstract}
Robots have the potential to enhance teaching of advanced computer science topics, making abstract concepts more tangible and interactive. In this paper, we present \textit{Timmy}--a GoPiGo robot augmented with projections to demonstrate shortest path algorithms in an interactive learning environment. We integrated a JavaScript-based application that is projected around the robot, which allows users to construct graphs and visualise three different shortest path algorithms with colour-coded edges and vertices. Animated graph exploration and traversal are augmented by robot movements. To evaluate \textit{Timmy}, we conducted two user studies. An initial study ($n=10$) to explore the feasibility of this type of teaching where participants were just observing both robot-synced and the on-screen-only visualisations. And a pilot study ($n=6$) where participants actively interacted with the system, constructed graphs and selected desired algorithms. In both studies we investigated the preferences towards the system and not the teaching outcome. Initial findings suggest that robots offer an engaging tool for teaching advanced algorithmic concepts, but highlight the need for further methodological refinements and larger-scale studies to fully evaluate their effectiveness.

\end{abstract}


\maketitle

\section{Introduction and Background}
\par The field of education has always been dynamic, adapting to the evolving needs of learners. Historically, teaching relied on tools such as chalkboards, textbooks, and rote memorisation. In recent decades, the teaching methods have transformed significantly with the rise of digital tools and resources, which also influenced educational strategies. The adoption of digital tools and resources has expanded the ways knowledge can be delivered and accessed. Today, educators educators emphasise the importance of designing interactive and immersive learning experiences~\cite{de2010learning, wagner2021creating}, exploring various approaches to enhance learning effectiveness, with a focus on promoting student engagement and collaboration. Comprehensive analyses, such as~\cite{TraditonalVSModern}, have sought to identify the most efficient teaching methods. 

\par Innovative teaching methods have incorporated robotics as an educational tool~\cite{reich2016robots}. With ready available kits, robotics offers accessible opportunities for learners to engage in programming and problem-solving activities. One such options is the GoPiGo Education Kit~\cite{gopigo2024}, which has been used in several educational contexts and topics~\cite{persson2016pedagogo, edwards2018not, cai2023exploring}. Building on the success of these prior works, our research investigates the potential of the GoPiGo robot in teaching advanced computer science topics, such as shortest path algorithms~\cite{gallo1988shortest}. Specifically, we assess students’ preferences for using the GoPiGo robot as an educational tool over more traditional methods, such as textbooks and screen presentations. To this end, we developed a learning environment where a robot is synced with on-screen visualisations when showing and teaching solutions to graph problems involving shortest path algorithms. We compared these robot-synced gestures with screen-only conditions to see which approach is preferred by students. The study seeks to contribute to a broader understanding of how robots augmented with visualisations can be integrated into educational practices.

\begin{figure}[h]
    \centering
    \includegraphics[width=0.95\linewidth]{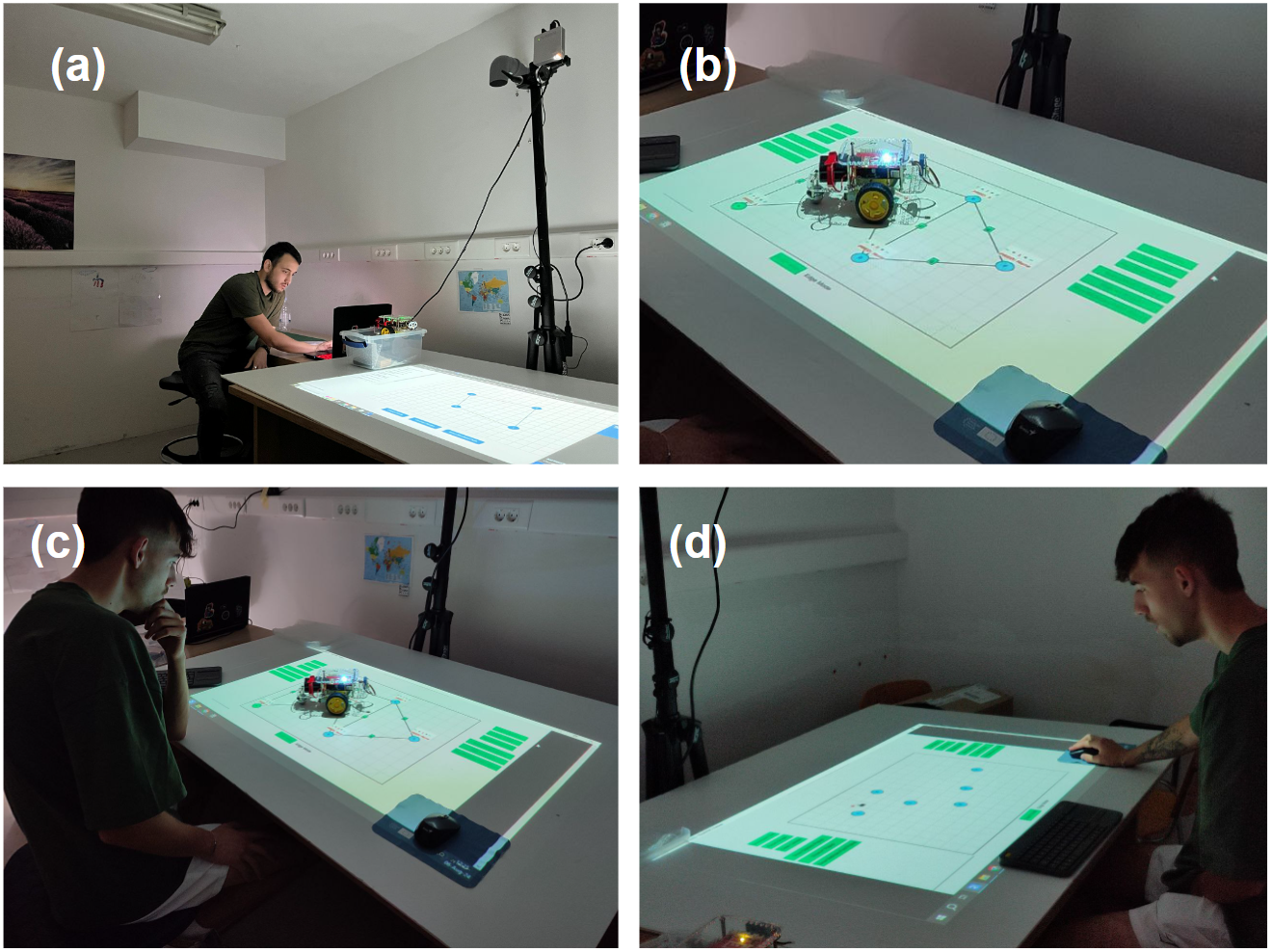}
    \caption{(a) Full setup of the project. (b) The robot in the projected environment. (c) Participant observing the robot in the robot condition. (d) Participant interacting with the application.}
    \label{fig:GroupedImage}
\end{figure}

\section{Timmy and the Learning environment: Design and Features}
\subsection{Overview}

\par
\begin{figure}[t]
    \centering
     \includegraphics[width=0.95\linewidth]{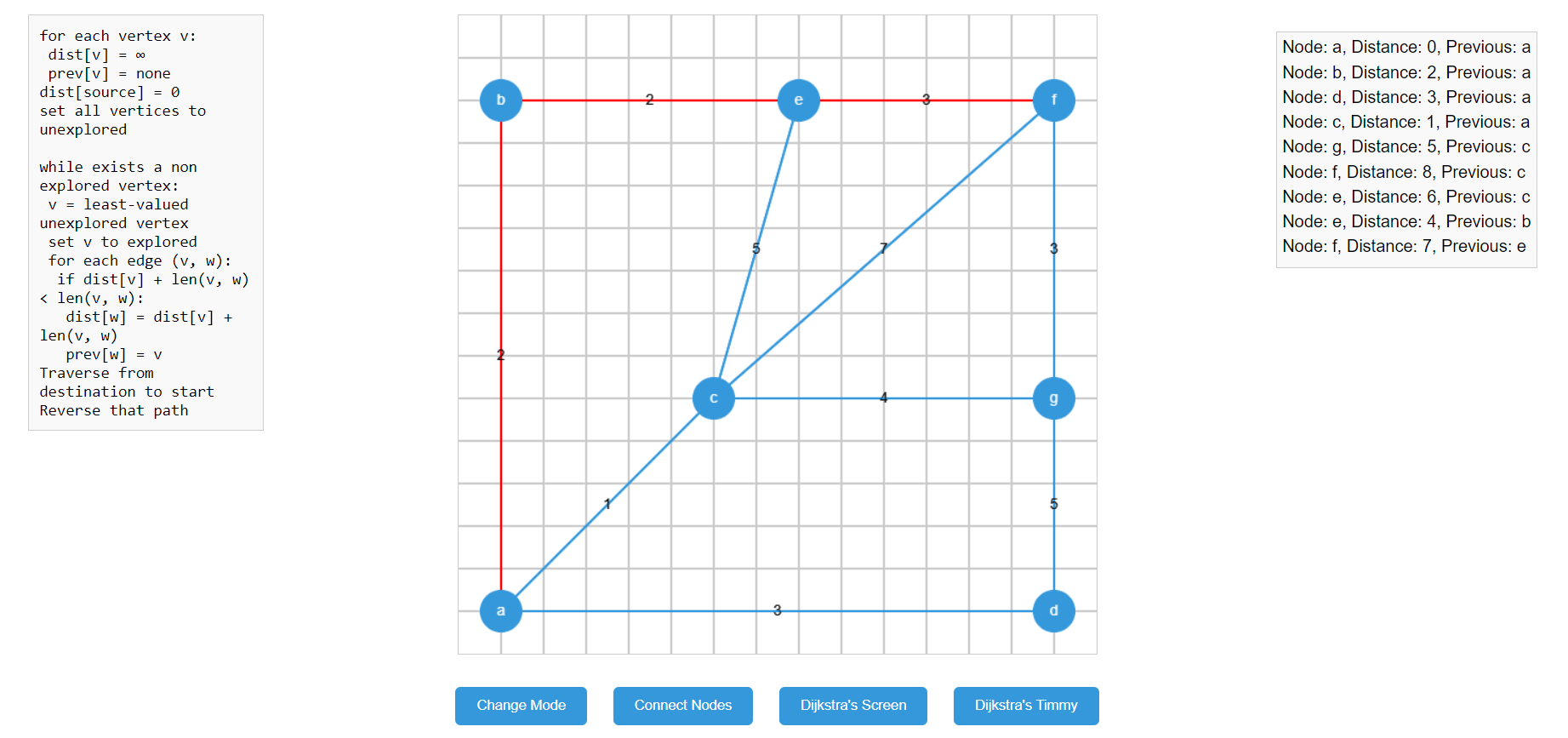} 
    \caption{The view of the application interface. On the left panel is the pseudocode. In the middle is the main canvas with the graph showing the vertices, edges and shortest path. On the right panel we see the distance and predecessor updates.}
    \label{fig:initial_interface}
\end{figure}

 The learning environment consists of two primary components (i) the robot \texttt{Timmy} and the (ii)~synchronised application. A projector is positioned above the setup and displays the application on the surface where \texttt{Timmy} can move. The application displays elements (nodes and vertices) that are animated in correct sequence and synced with the robot's movements, enabling users to visually understand the workings of the shortest path algorithms. \texttt{Timmy} is a \texttt{GoPiGo3} robot and runs on \texttt{GoPiGoOS} $3.0.3$. It is connected to the application via its Network Mode using RealVNC player\footnote{https://www.realvnc.com/en/connect/download/viewer/}.  The application was built using JavaScript and is run in a web browser. The setup of the system and the learning environment are shown in \autoref{fig:GroupedImage}. The application interface, depicted in~\autoref{fig:initial_interface}, includes (i) the pseudo code panel on the left, (ii) updates for distances and predecessors panel on the right, and (iii) a pre-drawn graph shown at the centre on the main canvas. 

\subsection{Main canvas}

\par The main canvas allows users to draw graphs to solve common shortest path algorithm problems (as can be seen in \autoref{fig:GroupedImage} (d)). It offers two drawing modes: (i) vertex mode for adding, deleting, and moving vertices, and (ii) edge mode, for adding, editing, and deleting edges. Once the graph is drawn, users can select one of the three shortest path algorithms to visualise: Dijkstra’s~\cite{dijkstra2022note}, A* (A star)~\cite{hart1968formal}, or Bellman-Ford~\cite{bellman1958routing}. They can also view the pseudo-code for each algorithm by clicking the respective button, either before or after the algorithm's execution. 

\par Each algorithm can be executed in two modes: on-screen-only mode or robot-synced mode. In on-screen-only mode, the graph exploration is visualised through sequenced animations of green edges. The shortest path is then highlighted  in red. In robot-synced mode, edge colouring is omitted, and instead, \texttt{Timmy} physically traverses the path. Users witness the robot's actual movement, simulating a real-world traversal of the shortest path. The application communicates with \texttt{Timmy} over a network, sending sequences and coordinates that the robot translates into physical movements, replacing the animated green line from the on-screen mode. This allows users to see the shortest path through the robot's physical movement. To ensure consistency across varying projection sizes, the grid's square dimensions were measured. The length of one side of a grid cell is multiplied with the unit values used in \texttt{Timmy}’s movement functions, ensuring \texttt{Timmy}’s movements remained consistent, regardless of the projection’s size.

\subsection{Synchronising Timmy's movements}

\par Since \texttt{Timmy} cannot directly ``see'' the graph, the graph data is sent to the robot via text files generated by the graph application. The file contains a list of $(x, y)$ coordinate pairs for vertices and an adjacency list for edges. \texttt{Timmy} reads the file name and responds accordingly: (i) Select a vertex: turn and move to the specified vertex; (ii) Poke a vertex: move to the vertex, then reverse back to the current position; (iii) Traverse the shortest path: move sequentially from vertex to vertex along the shortest path; (iv) Celebrate: spin a set number of times at the final vertex and blink the lights three times.

\par To ensure the robot moves accurately, its position and orientation are continuously tracked and stored in a \texttt{JSON} file, which is updated whenever a new graph file is downloaded. This process is handled by a Bash script, \texttt{watcher.sh}. To ensure smooth integration between the robot and the application, the time between file downloads and the next visualisation step is based on how long the robot takes to complete its actions. This timing is calculated by measuring the robot’s movement and turn rates, with extra time added for file processing. Since the robot can only perform one action at a time, precise synchronisation was crucial for ensuring seamless operation.

\section{Initial Study}

\par 
The initial study was conducted in the early implementation stage with only one algorithm implemented--Dijkstra's algorithm. At this stage we wanted to assess the feasibility of the proof of concept. 

\subsection{Protocol}

\par We recruited university students ($n=10$, aged 18 to 25) with prior knowledge in graph algorithms. The study employed a within-subject design, allowing participants to experience both the on-screen-only and robot-synced modes as the conditions in a randomized, counter-balanced order. After signing the consent forms, participants were introduced to the conditions, where they either observed the robot demonstrating the algorithm or watched its animation on the screen. Participants passively-observed the execution of the algorithm on the pre-drawn graph and were given time to review before continuing. 

\par After each condition, participants' cognitive load was assessed using the NASA-TLX questionnaire. They were then given a similar graph layout but with different weights and tasked with solving for the shortest path on paper.  The sequence was repeated for the other condition. Once participants had experienced both conditions, they completed a questionnaire to indicate their preferences:

\begin{enumerate}[noitemsep,label=Q\arabic*:]
 \item Which condition did you like the most? (Screen/Robot/Neither)
 \item Which condition do you think is better for learning shortest path algorithms?  (Screen/Robot/Neither)
 \item Which condition was easier for you to handle? (Screen/Robot/Neither)
 \item Which condition would you recommend to your colleagues? (Screen/Robot/Both/Neither)
 \item Would you like your teacher to use any of the conditions in class? (Screen/Robot/Both/Neither)
 \item Open Comments.
\end{enumerate}

\subsection{Findings}
\subsubsection{Cognitive Load across both conditions}
\begin{figure}[!t]
    \centering
    \begin{subfigure}{0.49\textwidth}
        \centering
        \includegraphics[width=\textwidth]{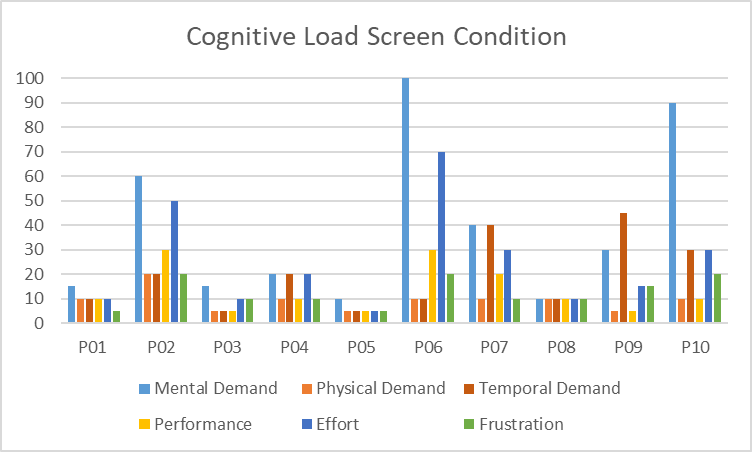}
        \caption{Cognitive load of the screen condition.}
        \label{fig:cognitiveLoadScreen}
    \end{subfigure}
    \begin{subfigure}{0.49\textwidth}
        \centering
        \includegraphics[width=\textwidth]{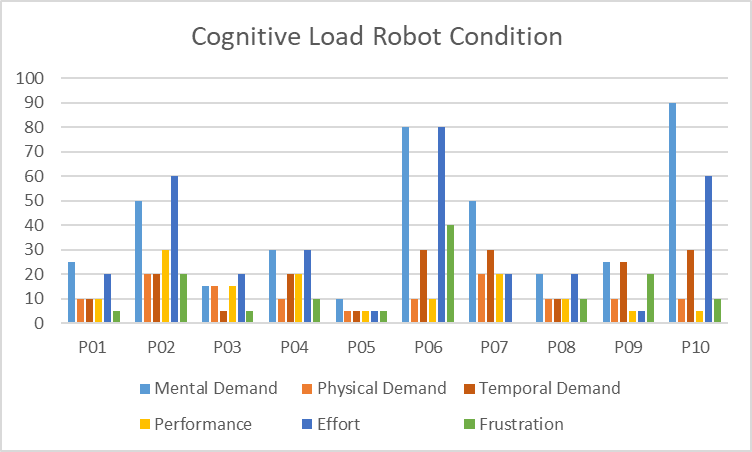}
        \caption{Cognitive load of the robot condition.}
        \label{fig:cognitiveLoadRobot}
    \end{subfigure}
    \caption{Cognitive loads of both conditions. The values are shown per participant and are not aggregated to show the varying levels across each dimension.}
    \label{fig:cognitiveLoads}
\end{figure}

\par \autoref{fig:cognitiveLoads} shows the subjective cognitive load values. Participants reported slightly higher mental demand for the on-screen-only condition compared to the robot-synced condition. Overall, the cognitive load across both conditions was slightly above 20\%. This indicates that while the cognitive load is manageable, there is room for improvement and a risk that increasing it further could overwhelm participants.  

\subsection{Participant Preference Findings}

\par All participants (I\_P01 to I\_P10) unanimously ($100\%$) preferred the robot-synced mode. All 10 also felt that learning with the robot was more effective at demonstrating shortest path algorithms. Interestingly, opinions were mixed regarding which condition was easier to handle, with 6 out of 10 participants selected robot-synced while 4 selected the on-screen-only. Despite this, participants indicated they would recommend both interfaces to their peers and expressed interest in having robot-based methods incorporated into the classroom instructions. While the robot-synced mode was the preferred choice, most participants agreed that both conditions offer unique advantages, highlighting the complementary benefits of each condition. 

\par The findings revealed promising potential for further development and 
highlighted the importance of providing diverse learning options to fit individual preferences and learning styles. 

\section{Pilot Study}

\par 
The pilot was conducted with the prototype fully implemented. 

\subsection{Protocol}

\par We recruited 6 participants, all university students ($n=10$, aged 18 to 25) with prior knowledge in graph algorithms. None of them participated in the initial study. Participants were introduced to the graph-drawing application and given time to familiarise with it. They were tasked with drawing a 6-vertex graph to ensure consistent difficulty. Once the graph was drawn, participants proceeded with the learning process. This study also followed a within-subject design, with $n=3$ participants per condition, and the conditions were counterbalanced. Participants observed and engaged with each condition in the same way as in the Initial study. 

\par Participants observed all three algorithms following the order of Dijkstra, A* and then Bellman-Ford for both conditions. 
To select an algorithm, users clicked the corresponding button. 
In the on-screen-only condition, the animation played upon algorithm selection. In the robot-synced condition, users were first given additional instructions on where to place the robot before observing its movement. This process was repeated each time they chose an algorithm. 
After completing each condition, participants filled out the System Usability Score (SUS) questionnaire. Once both conditions were finished, participants were asked about their preferences. A question was added to the questionnaire inquiring which of the two conditions would be more suitable for teaching primary school children (for teaching less complex concepts). 

\subsection{Findings}
\subsubsection{Reported System Usability}

\par SUS scores indicate that both the on-screen-only (mean: 80.42, std:~8.34) and the robot-synced (mean: 72.92, std: 6.02) conditions have similar usability scores, both considered above average. However, the on-screen-only condition was rated higher than the robot-synced condition. We suggest that this difference arises because, in the on-screen-only condition, participants only had to passively-observe the animations, whereas in the robot-synced condition, they needed to carefully position the robot. This aligns with observations in the initial study, where users did not have to handle the robot and simply observed it during its execution. 

\subsubsection{Participant Preference Findings}

\par Four out of six (66\%) participants preferred the robot-synced over the on-screen-only condition. Participant P\_P06 explained that their preference for the on-screen-only condition was due to the speed at which information was presented. They noted that the on-screen-only condition was significantly faster than the speed of how the information was presented in the robot-synced condition. This helped maintain their focus better.

\par Participants P\_P01 and P\_P02 who preferred the robot, found it visually engaging and felt it made the algorithm easier to understand. They also believed it helped maintain their attention, contrasting the claim of P\_P06 from Q1. However, this was not the case for participant P\_P04, who found the robot too distracting and preferred a more straightforward approach, stating, ``The more boring the better for me.'' Similar to P\_P06's comment in Q1, participant P\_P05 noted that the on-screen-only was more efficient time-wise. Meanwhile, participant P\_P06, despite favouring the on-screen-only, suggested that an ideal approach would be to combine both methods.

\par In Q3, participants were asked which method they believed would be better for teaching shortest path algorithms to primary school children. All participants chose the robot-synced condition. Many noted that the robot would be more engaging and enjoyable for children, making them more likely to stay focused and interested in the learning process. This is supported by the findings in~\cite{baxter2017robot} where robots were shown to promote learning in primary school settings. However, this suggests that while robots can be an engaging tool for teaching children, it does not necessarily mean they are suitable for teaching more advanced concepts such as shortest path algorithms. 

\par In Q4, participants were asked which method they found easier to handle. Four participants preferred the on-screen-only over the robot-synced one, while two believed that both were easy to handle. Participant P\_P01 explained their preference for the on-screen-only condition, noting that it required less handling ``Just clicking a button.'' 

\par All participants would recommend both the screen and the robot rather than picking a single choice. They mentioned that the choice should depend on individual preferences and tastes as both methods can be suited to different learning styles. All participants also answered ``Yes'' when asked whether they would like to have a robot as a tool for learning algorithms in general. Participant P\_P01 mentioned that the robot would make learning more interesting. Participant P\_P02 added that, as technology advances, there will likely be more tools like this.

\par In the open comments, participants offered suggestions for improving the user interface and overall experience. Participants P\_P01 and P\_P02 both recommended adding two buttons to enable switching between vertex-mode or edge-mode in the graph drawing component of the application, believing this would improve the graph drawing experience. Participant P\_P01 also suggested clarifying when the robot starts ``drawing'' the shortest path. Participant P\_P02 requested faster transitions on the on-screen-only method. Participant P\_P05 proposed using a smaller robot to reduce costs and make it more suitable for personal use. Participant P\_P06 mentioned difficulty maintaining attention while using the robot and suggested that colouring the lines could make it easier to track progress. They recommended using a grey colour for edges that are explored in the current iteration and marking edges that will not be revisited.

\par The pilot study findings indicate that user preference for educational methods depends significantly on user preference. Both the screen and robot are generally usable, but improving the robot’s usability to match or surpass the screen’s performance could improve its effectiveness. 
There is strong interest in using robots for educational purposes, suggesting that ongoing development and integration of robotic tools in learning environments could be highly beneficial.

\section{Discussion}

\par Many studies have explored the use of robots in education, primarily for teaching computer science and general subjects~\cite{fagin2003measuring, kay2012using, burbaite2013using, beer1999using}. Our research investigates whether robots have the potential to teach advanced computer science topics such as shortest path algorithms. While the \texttt{GoPiGo} is beginner-friendly, its limitations present challenges for advanced applications. For instance, the batteries deplete quickly, the VNC connection frequently experiences errors, requiring constant reconnection, and the weak processing power of the Raspberry Pi makes navigating the robot's OS desktop slow. This becomes particularly problematic when using the robot's desktop to run a browser. These limitations suggest that alternative robots may be better suited for educational projects.

\par The initial study shows that while participants initially preferred robots for learning, their preference shifted based on the task. In the initial study, where participants passively-observed the robot navigating a pre-built graph, the robot was favoured. However, in the pilot study, where participants had to build their own graphs and interact with the robot, they preferred the screen-based learning condition.

\par One limitation of our studies is also the small sample sizes (10 in the initial study and 6 in the pilot study). However, the results indicate potential for further exploration. Larger-scale studies are needed to assess the full potential of robots in the proposed educational context. Additionally, future research should include younger participants, as the current sample may not fully represent the target audience. Moreover, future research should explore the effectiveness of using such teaching methods by testing participants' understanding of the content being taught. Participants were recruited under the assumption that they already knew graph algorithms. We can consider some screening criteria to properly assess their knowledge before and after the experiments. 

\section{Conclusion}

\par This study explored the potential of using the \texttt{GoPiGo} robot to teach advanced computer science topics, specifically shortest path algorithms (Dijkstra’s, A*, and Bellman-Ford), by integrating a physical robot with a graph-drawing application, where the robot would traverse the graph drawn by users and demonstrate the procedure of the selected algorithm. Two user studies were conducted: (i) an initial study with participants just observing the procedure on the screen and with the robot traversing a graph for Dijkstra’s algorithm, and (ii) a pilot study where participants had to draw their own graph and interact with the system in order to observe the execution of all three algorithms on screen and with the robot. While small sample sizes (10 for the initial and 6 participants for the pilot study) limit the generalisability, results suggest that robots can maintain users' attention and show promise as educational tools. However, participants' preferences shifted depending on how they interacted with the robot, highlighting the importance of effective instructional design. Future research should involve larger, more diverse participant groups and focus on optimising instructional methods. 

\begin{acknowledgments}
  This research was funded by the Slovenian Research Agency, grant number P1-0383, P5-0433, IO-0035, J5-50155 and J7-50096. This work has also been supported by the research program CogniCom (0013103) at the University of Primorska.
\end{acknowledgments}

\bibliography{main.bib}




\end{document}